\documentclass{article}

\usepackage[final,nonatbib]{neurips_2019}

\usepackage[utf8]{inputenc} % allow utf-8 input
\usepackage[T1]{fontenc}    % use 8-bit T1 fonts
\usepackage{hyperref}       % hyperlinks
\usepackage{url}            % simple URL typesetting
\usepackage{booktabs}       % professional-quality tables
\usepackage{amsfonts}       % blackboard math symbols
\usepackage{nicefrac}       % compact symbols for 1/2, etc.
\usepackage{microtype}      % microtypography
\usepackage{amsmath}
\usepackage{amssymb}
\usepackage{graphicx}
\usepackage{bm}

\title{Patch Refinement - Localized 3D Object Detection}

\author{%
  Johannes Lehner \\ 
  Institute for Machine Learning\\
  Johannes Kepler University Linz\\
  \texttt{lehner@ml.jku.at} \\ 
  \And 
  Andreas Mitterecker \\
  Institute for Machine Learning\\
  Johannes Kepler University Linz\\
  \texttt{mitterecker@ml.jku.at} \\
  \And  
  Thomas Adler \\
  Institute for Machine Learning\\
  Johannes Kepler University Linz\\
  \texttt{adler@ml.jku.at} \\ 
  \And 
  Markus Hofmarcher \\
  Institute for Machine Learning\\
  Johannes Kepler University Linz\\
  \texttt{hofmarcher@ml.jku.at} \\
  \And
  Bernhard Nessler \\
  Institute for Machine Learning\\
  Johannes Kepler University Linz\\
  \texttt{nessler@ml.jku.at} \\
  \And
  Sepp Hochreiter \\
  Institute for Machine Learning\\
  Johannes Kepler University Linz\\
  \texttt{hochreit@ml.jku.at} \\
}

\begin{document}

\maketitle

% %%%%%%%%%%%%%%%%%%%%%%%%%%%%
% %%%%%%%%% ABSTRACT %%%%%%%%%
% %%%%%%%%%%%%%%%%%%%%%%%%%%%%
 \begin{abstract}
We introduce Patch Refinement a two-stage model for accurate 3D object detection and localization from point cloud data. Patch Refinement is composed of two independently trained Voxelnet-based networks, a Region Proposal Network (RPN) and a Local Refinement Network (LRN). We decompose the detection task into a preliminary Bird’s Eye View (BEV) detection step and a local 3D detection step. Based on the proposed BEV locations by the RPN, we extract small point cloud subsets ("patches"), which are then processed by the LRN, which is less limited by memory constraints due to the small area of each patch. Therefore, we can apply encoding with a higher voxel resolution locally. The independence of the LRN enables the use of additional augmentation techniques and allows for an efficient, regression focused training as it uses only a small fraction of each scene. 
Evaluated on the KITTI 3D object detection benchmark, our submission from January 28, 2019, outperformed all previous entries on all three difficulties of the class car, using only 50\,\% of the available training data and only LiDAR information.

\end{abstract}

% %%%%%%%%%%%%%%%%%%%%%%%%%%%%%%%%
% %%%%%%%%% Introduction %%%%%%%%% 
% %%%%%%%%%%%%%%%%%%%%%%%%%%%%%%%%
\section{Introduction}
Object detection and localization is one of the key challenges in autonomous driving \cite{Geiger2012kitti} and robotics \cite{Wolf2016robotics}. Compared to the performance of 2D object detection in images the results in 3D object detection lag behind considerably, mainly caused by increased difficulty of the localization task. Instead of fitting axis-aligned rectangles around the part of objects that is visible in the image plane, the main challenge in 3D detection and localization is the amodal oriented 3D bounding box prediction in 3D space, which includes the occluded or truncated parts of objects.

Despite the recent advancement of image-only 3D detectors \cite{wangcvpr2019,licvpr2019} point cloud information from LiDAR sensors is a requirement to achieve highly accurate 3D localization. Unfortunatly, LiDAR data is represented as an unordered, sparse set of points in a continuous 3D space and can not be directly processed by standard convolutional neural networks (CNNs).
Deep CNNs operating on multiple discretized 2D Bird's Eye View (BEV) maps achieved early success in both 3D \cite{Chen2017multiview,Ku2018avod} and BEV \cite{Yang2018pixor} detection. Simply discretizing point clouds is inevitably linked to a loss of information.
Current state-of-the-art 3D object detection is largely based on the seminal work PointNet \cite{Qi2017pointnet}. Pointnets are used in combination with 2D image detectors to perform 3D localization on point cloud subsets \cite{Qi2018fpointnet,Shin2018roar}. Two-stage approaches \cite{Shi2019pointrcnn} use PointNets in an initial segmentation stage and a subsequent localization stage.
Voxelnet \cite{Zhou2018voxel} introduces Voxel Feature Encoding (VFE) layers which utilize PointNet to learn an embedding of local geometry within voxels, which can then be processed by 3D and 2D convolutional stages. Others \cite{Yan2018second,Lang2019pointpillars} add further improvements to Voxelnet.

\paragraph{Motivation} Voxelnet is a single-stage model, as such it applies VFE-encoding with a uniform resolution on the whole scene, although such a resolution is only necessary at locations that contain objects. As a consequence, it is severely limited by memory constraints, especially during training, where only a batch size of 2 can be processed by a GPU with 11 GB of memory. Given the typical sparseness of objects within a LiDAR scene, only a few subsets contain viable information to train regression \cite{Engelcke2017vote3deep}. But in the case of Voxelnet Batch Normalization layers \cite{Ioffe2015batchnorm} break this local independence.

\paragraph{Patch Refinement} To be able to train a small detector focused on 3D bounding box regression, we decompose the task into a preliminary BEV detection step and a local 3D detection step, similar to the two-stage approach of R-CNN \cite{Girshick2014rich}. Object sizes are bound and unaffected by the distance to the sensor. We construct a Local Refinement Network (LRN) that operates on small subsets of points within a fixed-sized cuboid, which we term ''patches''. The RPN does not have to perform warping and independence of the LRN can be achieved by training with some noise to account for proposed locations that are slightly offset. We favor an independent approach because it enables additional augmentation options, the higher resolution features have to be calculated in any case and it allows us to evaluate the regression ability of the LRN without the influence of an RPN.

\begin{figure}[ht!]
    \centering
    \includegraphics[width=0.55 \textwidth]{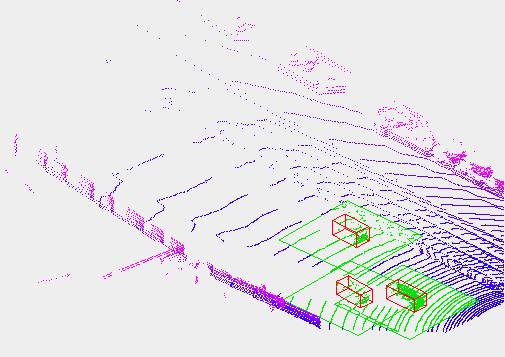} \includegraphics[width=0.39 \textwidth]{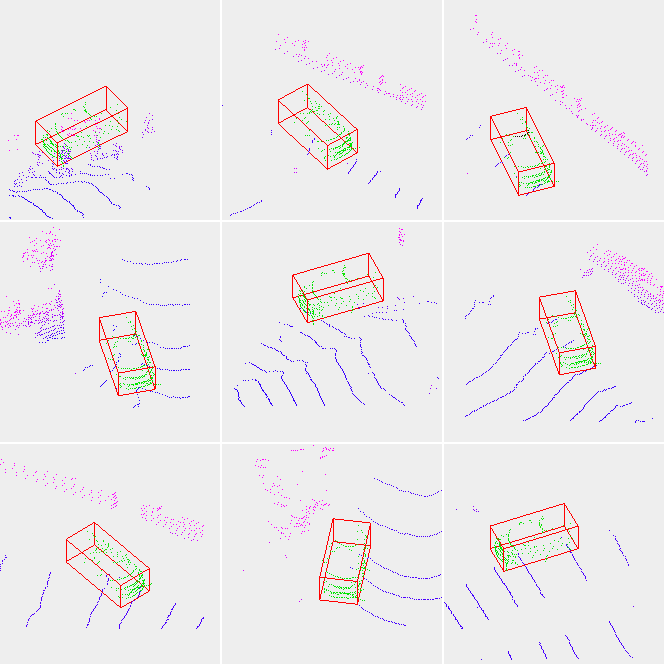}
    \caption{ Left: A visualization of extracted subsets (``patches'') - highlighted in green with red bounding boxes - which serve as input to our Local Refinement Network (LRN). A patch contains the points within a cuboid centered at the object. Right: patches constructed to train the LRN, the same object is placed on different surfaces and occurs in different orientations.}
    \label{fig:patches_vis}
\end{figure}

Our work comprises the following key contributions: 
\begin{itemize}
\item We demonstrate that it is a viable approach to decompose the 3D object detection task in autonomous driving scenarios into a preliminary BEV detection followed by a local 3D detection via two independently trained networks.
\item We show that it takes only a few simple modifications to utilize the Voxelnet \cite{Zhou2018voxel} architecture either as an efficient Region Proposal Network or as a Local Refinement Network that is capable to perform highly accurate 3D bounding box regression and is not limited to fixed input space. 
\item We report the beneficial effect of adding the corner bounding box parametrization of AVOD \cite{Ku2018avod} as auxiliary regression targets, even without applying the proposed decoding algorithm.
\end{itemize}

% %%%%%%%%%%%%%%%%%%%%%%%%%%%%%%%%
% %%%%%%%%% RELATED WORK %%%%%%%%%
% %%%%%%%%%%%%%%%%%%%%%%%%%%%%%%%%
\section{Related Work}

MV3D \cite{Chen2017multiview} and AVOD \cite{Ku2018avod} are two-stage models that fuse image and point cloud information and perform regression on the resulting 2D BEV feature maps, projections of point clouds and camera information. While this enables the use of 2D CNNs, these approaches cannot capture the full geometric complexity of the 3D input scene, due to information loss caused by discretization. 

Frustum PointNets \cite{Qi2018fpointnet} projects the proposals of an image-based 2D detector onto the point cloud. The resulting frustum is then further processed by a sequence of PointNets.
It demonstrates that accurate amodal bounding box prediction can be performed without context information, which is actively removed by a segmentation PointNet. Noticeably, detection scores are calculated without taking the LiDAR representation into account.

Voxelnet \cite{Zhou2018voxel} applies a 3D grid to divide the input space into voxels. Followed by a  sparse voxel-wise input encoding via a sequence of PointNet-based VFE layers. 
This enables the network to learn structures within voxels and to embed the point cloud into a structured representation while retaining the most important geometric information in the data. Which are subsequently processed by 3D and 2D CNNs.
Both our RPN and LRN are based on Voxelnet, modified to better accomplish their respective tasks.

SECOND \cite{Yan2018second} modifies Voxelnet by replacing the costly dense 3D convolutions with efficient sparse convolutions, reducing both run-time and memory consumption considerably.
Furthermore, they propose ground truth sampling, an augmentation technique that populates training scenes with additional objects from other scenes.
Besides speeding-up training by increasing the average number of objects per frame, this augmentation provides strong protection against overfitting on context information \cite{Shi2019pointrcnn,Lang2019pointpillars}.

PointPillars \cite{Lang2019pointpillars} proposes a VoxelNet-based model without 3D convolutional middle layers. Instead, they apply a voxel grid with a vertical resolution of one and the encoding of the vertical information is solely performed within the VFE-layers. Further optimized towards speed, the model achieves the highest frame rates within the pool of current 3D object detection models. Our RPN follows a similar design, but we use a vertical resolution of two and concatenation.

PointRCNN \cite{Shi2019pointrcnn} is a two-stage approach utilizing PointNets, that introduces a novel LiDAR-only bottom-up 3D proposal generation first stage, followed by a second stage to refine predictions.
Similar to our model it follows the R-CNN approach and pools the relevant subset of the input point cloud for each proposal.
Unlike our LRN, the second stage of Point R-CNN reuses higher-level features and relies on the RPN to transform the proposals into a canonical representation.

% %%%%%%%%%%%%%%%%%%%%%%%%%%%%%%%%
% %%%%%%%%% Method %%%%%%%%%%%%%%%
% %%%%%%%%%%%%%%%%%%%%%%%%%%%%%%%%
\section{Method} \label{sec:method}

\begin{figure*}[ht!]
\centering
\includegraphics[width=0.95\textwidth]{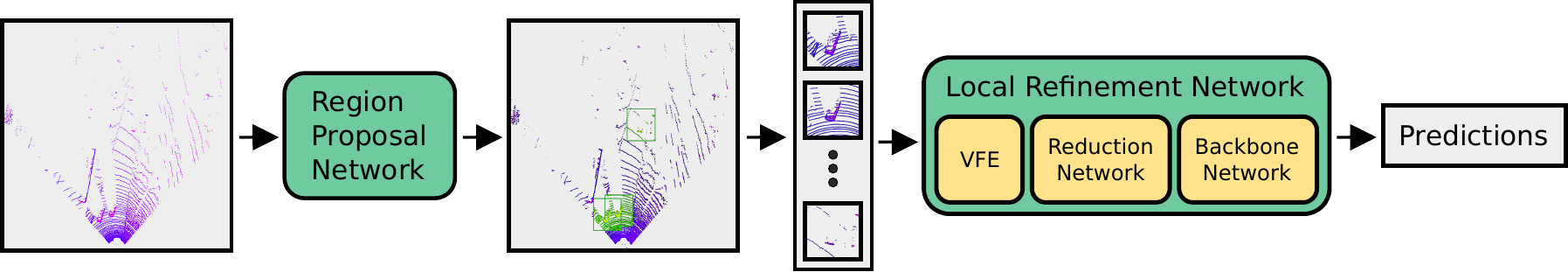} 
\caption{The region proposal network (RPN) identifies the  locations of objects in Bird's Eye View (BEV). Based on these proposed locations subsets are extracted from the point cloud and processed by the local refinement network (LRN).}
\label{fig:architecture}
\end{figure*}

Figure~\ref{fig:architecture} depicts the inference procedure, which follows the original R-CNN \cite{Girshick2014rich} approach. 

\subsection{Local Refinement Network} \label{sec:lrn}
We follow the Voxelnet approach and apply a 3D voxel grid to the input, grouping points to voxels. 
This is followed by a VFE stage and then by a reduction step from 3D to 2D BEV feature maps. 
These are then processed by our 2D convolutional backbone network. 

\paragraph{Grouping Points into Voxels}
We use the efficient sparse encoding algorithm of Voxelnet that processes only non-empty voxels. Although we train only on small regions of the input scene, we preserve absolute (global) point coordinates. This way our model can learn that objects farther from the sensor are typically represented with fewer measurements than nearby objects. 

\paragraph{Voxel Feature Encoding}
While the grouping algorithm of Voxelnet processes only non-empty voxels, the input to the VFE layers consists of dense tensors with a large proportion of padded zeros (roughly 90\,\% with the default setting of at most 35 points per voxel). 
As it has a regularizing effect on the running mean and variance this zero-padding alleviates the use of Batch Normalization (BN) \cite{Ioffe2015batchnorm}. 
We relinquish the use of zero padding and remove the BN in our VFE-stage. 
Instead, we apply per-sample normalization. Overall, our modified VFE layer requires less memory while also increasing the predictive performance of our model.

\paragraph{Reduction Network}
The activation tensor resulting from the VFE stage is still three dimensional.
Via multiple 3D-convolutional layers, the vertical resolution of the activation tensor is reduced to one resulting in 2D feature maps. The result is a 2D representation in BEV with vertical information encoded locally.

\begin{figure}[ht]
  \centering
  \includegraphics[width=0.45 \textwidth]{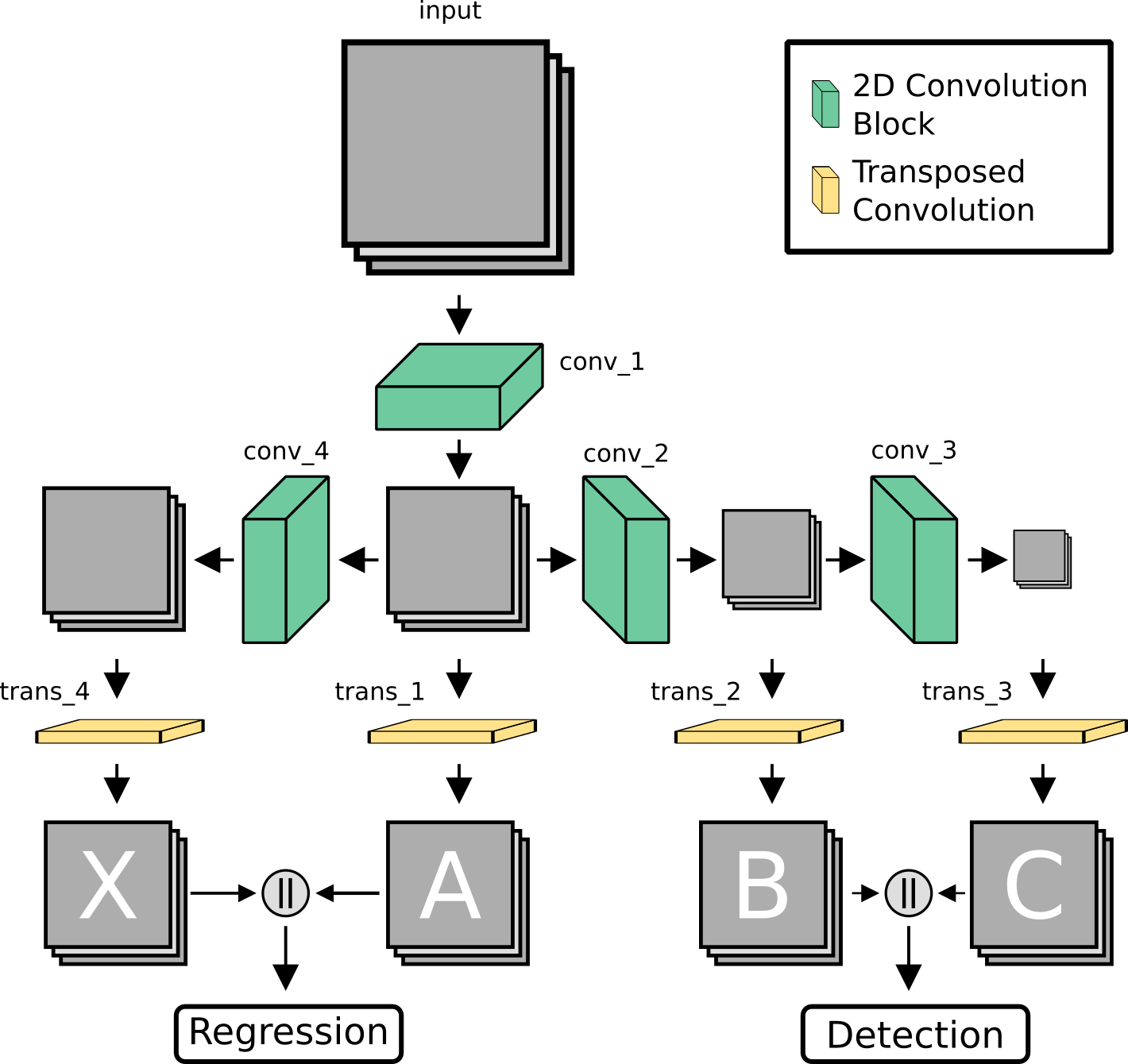}
  \caption{Architecture of the backbone network. The network is designed in a way that regression is performed on feature maps of higher resolution and limited receptive fields, while detection is performed on higher level feature maps with a larger receptive field. Before regression, we concatenate feature maps $A$ and $X$ and before detection we concatenate $B$ and $C$.}
  \label{fig:backbone}
\end{figure}

\paragraph{Backbone Network}
Figure~\ref{fig:backbone} depicts the architecture of our backbone network.
It uses multiple blocks of 2D convolutional layers to generate intermediate feature maps of different resolutions. We then apply transposed convolutional layers to get feature maps with equal dimensions before the output layers, where feature maps are concatenated and combined with $1 \times 1$ convolutions at each anchor position.
The resulting feature maps are labeled $A, B, C,$ and $X$ in Figure~\ref{fig:backbone}. 
For the regression head, the receptive field of $A$ has been chosen to cover the majority of objects while the receptive field of $X$ is slightly larger to cover outliers.
Restricting the receptive field of the regression head reduces distractions from the environment when predicting exact bounding boxes.
In contrast, we perform the detection based on higher-level feature maps that have a larger receptive field.

\paragraph{Loss}
We use a variant of residual box parametrization with a direction classifier \cite{Yan2018second} and sine and cosine encoding for orientation. 
From the vector $\bm{g} = (x,y,z,l,h,w,\theta)^\top$ representing the three box-center coordinates $x, y, z$, height $h$, width $w$, length $l$ and yaw $\theta$ around the $y$-axis of an oriented 3D bounding box, we calculate 
residual regression targets $\bm{u} = (\Delta x, \Delta y, \Delta z_{b}, \Delta z_{t}, \Delta w, \Delta h, \Delta l, \Delta \eta, \Delta \zeta)^\top$. In $\Delta z_{b}$, $\Delta z_{t}$, the subscripts $b$ and $t$ denote bottom and top respectively, between a ground truth vector $\bm{g}$ and an anchor vector $\bm{a}$ from the same vector space as described in Equation~\eqref{eqn:box_regression}:

\begin{equation}
\begin{aligned}
  \Delta x &= \frac{x_g - x_a}{\sqrt{l^2+w^2}} & 
  \Delta z_{b} &= z_g - \frac{h_g}{2} - z_a + \frac{h_a}{2} \\
  \Delta y &= \frac{y_g - y_a}{\sqrt{l^2+w^2}} & 
  \Delta z_{t} &= z_g + \frac{h_g}{2} - z_a - \frac{h_a}{2} \\
  \Delta l &= \log\frac{l_g}{l_a} &  \Delta w &= \log\frac{w_g}{w_a} & \Delta h &= \log\frac{h_g}{h_a} \\
  \Delta \zeta &= \lvert\cos(\theta_g - \theta_a)\rvert &
  \Delta \eta &= \sin{(\theta_g - \theta_a)} &
  \delta &= 
 \begin{cases}
     1 \enspace \text{if} \cos{(\theta_g - \theta_a)} > 0 \\ 
         0 \enspace \text{otherwise}
     \end{cases}
 \end{aligned},
 \label{eqn:box_regression}
 \end{equation}
 where $h_g$  denotes the component $h$ of the vector $\bm{g}$ (the notation 
 for the other components follows respectively).

Additionally, we use the box parametrization of AVOD \cite{Ku2018avod} as auxiliary regression targets. We transform ground truth boxes and anchor boxes into a 2D corner representation in BEV $(c,d)$ for the $x$- and $y$-coordinates of each corner resulting in eight variables $\bm{f} = (c_1, c_2, c_3, c_4, d_1, d_2, d_3, d_4)^\top$, where the subscripts enumerate corners. Then we calculate the targets $\bm{v} = \bm{f}_g - \bm{f}_a$.
While we do not use these additional regression parameters during inference, the added training task improves performance considerably.
We use axis-aligned 2D Intersection over Union (IoU) in BEV as a similarity measure between the ground truth boxes rotated to the nearest axis and the corresponding anchor type. 
For detection, positive anchors have to surpass an IoU threshold of $0.6$, while anchors with an IoU below $0.45$ are treated as negatives. 
For regression, the positive threshold is set to $0.45$. 
Our default choice for training the detection head is the balanced sampling of $N_{total}$ anchors with a ratio of 3 to 1 of negative and positive anchors. 
For detection and direction we use the binary cross-entropy loss function, denoted $L_{\text{cls}}$ and for the regression targets we use the smooth L1 loss (also known as Huber loss), denoted $L_{\text{reg}}$. Balancing is achieved via the hyperparameters $\alpha,\beta,\gamma$, with the default values of 1, 1 and 2 respectively.

Let $i$ and $j$ denote the sampled positive and negative detection anchors and let $p$ denote the sigmoid activation of the classification output. Further, let $k$ denote the positive regression anchors. $N_{pos\_reg}$ is the total number of positive regression anchors. The overall loss function is given as in Equation \eqref{eqn:loss_overall}.

\begin{equation}
\begin{aligned}
 L &= \alpha\frac{1}{N_{\text{\scriptsize total}}} \sum\limits_{i} L_{\text{\scriptsize cls}}(p^{\text{\scriptsize pos\_cls}}_i, 1) 
     + \beta\frac{1}{N_{\text{\scriptsize total}}} \sum\limits_{j} L_{\text{\scriptsize cls}}(p^{\text{\scriptsize neg\_cls}}_j, 0) \\
     &+ \gamma \frac{1}{N_{\text{\scriptsize pos\_reg}}} \sum\limits_{k} \Big(L_{\text{\scriptsize reg}}(\bm{u}_k, \bm{u}^*_k) 
    + L_{\text{\scriptsize{reg}}}(\bm{v}_k,\bm{v}^*_k) + L_{\text{\scriptsize cls}}(h_k,h_k^*)\Big)\\
 \end{aligned}
 \label{eqn:loss_overall}
 \end{equation}

\subsection{Patch Construction}\label{sec:augmentation}
The independence of the LRN enables us to construct patches from ground truth annotations.

\paragraph{Surface Sampling}
Inspired by the beneficial effect of ground truth sampling in SECOND, we try to achieve protection against overfitting on context information in a similar way. First, we create lists of objects and related surrounding areas (surfaces) in the training set. We then remove points within the slightly enlarged bounding boxes of the objects present in the respective surface. We then rotate each surface to align its center with the depth axis in front of the sensor car. Finally, we sort both object and surface lists based on the absolute distance to the sensor. 
During training, we combine objects with surfaces that appear in a similar distance to the sensor. To ensure that the object lines up with the surface, we look up the vertical coordinate of the original object and place the augmented object at the corresponding height. We apply surface sampling based on the difficulty levels easy, moderate and hard with the respective probabilities of $100\%$, $80\%$ and $60\%$.

\paragraph{Global \& Per-Object Augmentation}
We then proceed with standard augmentation techniques. Scaling and mirroring of the patch and the object individually.
Contrary to \cite{Zhou2018voxel,Yan2018second} we do not apply per-object rotation and vertical translation. As per-object rotation introduces self-occlusion artifacts and vertical translation creates unreasonable patches where objects appear below ground or fly above it. 
Especially in the context of self-driving, the training data for car objects is heavily biased, with a strong peak for parallel orientation to the ego-vehicle. A major benefit of working on a per-object basis without a fixed input space is the ability to use an augmentation for a full range of rotations in the forward view around the global $z$-axis by $[-\frac{\pi}{2},\frac{\pi}{2}]$. 
Therefore, every object is learned to be recognized in every possible orientation inside the patch, respecting the global position in which the so rotated object would happen to appear. Consequently, both the perpendicular and the parallel anchor types are trained equally well.

\paragraph{Random Cropping - LRN Detection Objective}
At this stage, we have an augmented object placed somewhere upon a surface. To achieve robustness against imperfect proposals by an actual RPN and to create a training task for the detection head, we sample noise from a circular area ($[-\pi,\pi]$ and $[0, 3]$ meters). Finally, we crop the patch at the BEV location that is determined by the object center and the offset from the sampled noise. Therefore, the objective of the detection head of the LRN will be to revert this offset and determine the correct object center within the small anchor grid. A task that is closely related to regression and designed to achieve a correlation between detection score and the ability to accurately regress an object.

\subsection{Region Proposal Network}
Our RPN is a slim version of the LRN network, the main source of simplification is the circumvention of the vertical reduction stage. Instead of multiple 3D convolutional layers, we use a vertical voxel resolution of two and concatenate the two resulting feature planes.
While this change reduces the 3D detection results considerably, the BEV detection results remain nearly unchanged. 

\subsection{Inference} \label{sec:inference}
During inference, we take the proposals of an RPN to extract patches. To increase the similarity with our training task, we remove points within the slightly enlarged bounding box of additional proposals falling within the patch. Similar to training we rotate the patch upon the depth axis (improvement of +0.15 AP).

 \section{Experiments}
 \paragraph{Data Set}
We evaluate our method on the KITTI 3D object detection benchmark \cite{Geiger2012kitti}, which provides samples of LiDAR point clouds as well as RGB camera images. 
The data set consists of 7,481 training samples and 7,518 test samples. 
Detections are evaluated on three difficulty levels (easy, moderate, hard). 
Average precision (AP) is used as the metric, successful detections for the class car require an IoU of at least 70,\%. 

\paragraph{Experimental Setup}
We subdivide the original training samples into a 3,712 samples train set and a 3,769 samples validation set as described in \cite{Chen2015objprop}, which we use for all our experiments and the submission to the test server. Furthermore, we use a non-maximum suppression threshold of 0.01. We train our model on a single 1080Ti GPU.

\subsection{Implementation Details}

\paragraph{Region Proposal Network} For our RPN we use the input space and BEV resolution of Voxelnet, a vertical resolution of 2, the described loss and a backbone configuration of ABC/AX. To compensate for the smaller receptive fields due to the circumvented 3D convolutions, we add 2 layers to the convolutional block 1. We first pre-train the RPN on patches with the local objective and only afterward train it for its final task as an RPN, simply by changing its input to whole point clouds and decreasing the learning rate by a factor of 0.1 and the batch size to 4. The RPN adapts to the new input within a few epochs only. Due to the increased imbalance between foreground and background objects, present when processing whole point clouds, we choose Focal Loss \cite{Lin2017focal} with default parameters $\alpha=0.25$ and $\gamma=2$ to train the detection head. 

\paragraph{Local Refinement Network}
For the width $x$ and depth $y$ dimensions, we chose voxel sizes of $0.15$ meters and patch sizes of $9.6$ meters. For the vertical dimension, we chose a voxel height of $\frac{4}{19}$ meters. In order to reduce the vertical dimension from 19 to 1, we use a sequence of 4 3D convolutional layers with a kernel-size of 3, vertical strides of (2,1,2,1), and no padding. To compensate for the smaller receptive fields due to the increased resolution we add 5 layers to the convolutional block 1 and 2 layers to convolutional block 2. The convolutional block related to the extended feature maps X  consists of 6 3x3 convolutions without initial down-sampling.
The voxel resolution is $64\times64\times19$ in the encoding stage. The regression and detection heads operate on a $32\times32$ anchor grid. 
We train with a batch size of 32 samples for 5 million samples with Adam Optimizer, an initial learning rate of $10^{-4}$ and after one million samples we multiply the learning rate by $0.8$ every $500,000$ samples. 

\subsection{Evaluation on the KITTI Test Set}
Table~\ref{tab:test_results} presents the results of our method on the KITTI test set using the Average Precision (AP) metric. 
Consuming only LiDAR data and 50\,\% of the training data, our submission outperformed all previous methods for 3D object detection on cars. Noticeably, on the easy difficulty, the effect of the local training objective of the detection head is most prominent. A comparison of the precision-recall curves provided by the KITTI benchmark showed that our model has a distinct advantage to better avoid high ranked false positive detections that do not pass the 70\,\% IoU threshold.

\begin{table*}[ht]
 	\small 
 	\begin{center}
 		\scalebox{0.975}{
 			\begin{tabular}{c|c||ccc|ccc}
 				\hline
 				{Method} & {Modality} & 			
				\multicolumn{3}{c|}{3D Object Detection} & \multicolumn{3}{c}{BEV Detection} \\
 				&&Easy & Moderate & Hard & Easy & Moderate & Hard\\
 				\hline\hline
 				MV3D \cite{Chen2017multiview} & RGB + LiDAR & 71.09 & 62.35 & 55.12 & 86.02 & 76.90 & 68.49 \\
 				Voxelnet \cite{Zhou2018voxel} & LiDAR & 77.47 & 65.11 & 57.73 & 89.35 & 79.26 & 77.39 \\
                F-PointNet* \cite{Qi2018fpointnet} & RGB + LiDAR & 81.20 & 70.39 & 62.19 & 88.70 & 84.00 & 75.33 \\	
 				AVOD-FPN* \cite{Ku2018avod} & RGB + LiDAR & 81.94 & 71.88 & 66.38 & 88.53 & 83.79 & 77.90 \\	
 				SECOND \cite{Yan2018second} & LiDAR & 83.13 & 73.66 & 66.20 & 88.07 & 79.37 & 77.95 \\
 				PointPillars* \cite{Lang2019pointpillars} & LiDAR & 79.05 & 74.99 & 68.30 & 88.35 & 86.10 & {\bf79.83}\\
 				PointRCNN \cite{Shi2019pointrcnn} & LiDAR & 85.94 & 75.76 & 68.32  & 89.47 & 85.68 & 79.10 \\
            	MMF* \cite{Liang2019CVPR} & RGB + LiDAR & 86.81 & 76.75 & 68.41  & 89.49 & {\bf87.47} & 79.10 \\
 				\hline 
 				Ours & LiDAR & {\bf87.87} & {\bf77.16} & {\bf68.91} & {\bf89.78} &  86.55 & 79.22 \\
 				\hline
 			\end{tabular}
 		}	
 	\end{center}
 	\caption{A performance comparison of 3D object detection methods on the KITTI 3D object detection benchmark and BEV benchmark for the class \emph{car}. Values are average precision (AP) scores on the official test set. Methods which use a custom or unspecified train-val split are marked with a `*'.}
 	\label{tab:test_results}
 	\vspace{-3mm}
 \end{table*}

\subsection{Ablation Studies on the KITTI Validation Set}

\begin{table}[ht]
\begin{center}
\begin{tabular}{l|c|c|c}
\hline
Modification & Easy & Moderate & Hard \\
\hline\hline
Patch Refinement & 89.55 & 79.04 & 78.10  \\
\hline\hline
original VFE & -0.60 & -0.69 & -0.85 \\
without auxiliary targets & -0.42 & -0.57 & -0.67 \\	
\hline
no surface sampling & -1.25 & -1.26 & -1.93 \\
reduced global rotation & -1.05 & -0.96 & -0.94 \\
\hline
backbone ABC/AX & -0.25 & -0.39 & -0.40\\
backbone C/AX & -0.20 & -0.19 & -0.06\\
backbone B/AX  & -0.24 & -0.32 & -0.35 \\
backbone BC/A & -0.54 & -0.40 & -0.26\\
backbone BC/AX red. & -0.06 & -0.03 & -0.15 \\
\hline
\end{tabular}
\end{center}
\caption{Ablation studies. 3D object detection scores, averaged over three LRN checkpoints on the validation set.}
\label{tab:ablations}
\end{table}

\paragraph{VFE-layers}
We trained our network with VFE-layers as proposed in Voxelnet, with zero-padding and BN. In this case, we observe a large performance drop overall difficulty levels when used in our model. We hypothesize that the reason is that typically
the batch statistics of those features are highly variable due to a low number of points and a varying input space. 
Additionally, we applied BN without zero-padding. This performs poorly both on whole scenes, as well as on patches.

\paragraph{Augmentation}
We analyze the effectiveness of our augmentation strategies: 
surface sampling and additional global rotation. 
Table~\ref{tab:ablations} shows two strong performance drops when
either of those is not employed. 
For surface sampling, this drop is expected, since we fully rely on the construction of artificial patches to avoid the model to overfit on context information.
The drop caused by reduced global rotation could be related to
an increased orientation bias. The data set comprises largely objects parallel to the sensor car and few objects in the perpendicular
orientation. As we rotate all objects
upon the depth axis and sample rotation only from $[\frac{-\pi}{4},\frac{\pi}{4}]$,
the number of perpendicular objects is further reduced. 
Overall, both augmentation strategies are vital components of the
training procedure. 

\paragraph{Auxiliary Regression Targets}
In our experiments without auxiliary regression targets, we observe slower, more unstable learning.
Additionally, with auxiliary regression targets, the model becomes more decisive in rejecting false positives and preserves the level of recall with fewer proposals. Table~\ref{tab:ablations} shows a decrease in performance if auxiliary regression targets are omitted. 

\paragraph{Backbone Modification}
We validate our architecture design against variants 
(see Figure~\ref{fig:backbone}), in which we modified
the connections of the regression head and detection head to
the feature maps A, B, C, and X. In the standard variant,
the detection head is connected to B and C, while the
regression head is connected to A and X. We denote this backbone as BC/AX. 
As a first variant, the backbone ABC/AX uses an additional feature map
for detection, which leads to earlier overfitting of the detection head. 
A comparison of the backbone variant C/AX and backbone variant B/AX using only one feature
map for detection, suggests that the higher level map C is of greater importance.
The backbone variant BC/A relinquishes the use of the additional regression map, which
has a negative effect on the performance on easy and moderate difficulty levels.  
The backbone variant ``BC/AX red.''  uses fewer layers in the convolutional
blocks 1 and 2, which achieves almost identical performance. 
Table~\ref{tab:ablations} shows the decrease in
performance for the different backbone variants.

\paragraph{Patch Extraction}
We study the influence of additional objects present in a patch. We calculate thresholds based on the percentile of the detection scores over the validation set. We then remove additional objects only if their respective detection score exceeds a given threshold.
We observe that removing those objects where the RPN is most confident is of greater importance and that the detection of objects of difficulty hard is affected the most from additional objects within a patch. We conclude that the effect is caused by distraction effects of additional objects with distinct features. 

\subsection{Additional Experiments}
\paragraph{Refinement of Other Models}
We study how our LRN performs when the region proposals are provided by other detection models. 
To this end, we construct validation patches based on the predictions generated by two other models.
Our experiments show only marginal differences between our RPN and two state-of-the-art detection models, namely SECOND v1.5 and 
PointRCNN (see Table~\ref{tab:different_RPNs}). The results underline that the regression capability of the RPN is of low importance.
Additionally, we compare our RPN to proposals created from ground truth labels. The gap increases with difficulty and suggests further improvement can be achieved via an RPN that is more capable to distinguish objects described by a low amount of LiDAR points, e.g. a fusion-based RPN.
\begin{table}
\begin{center}

 \begin{tabular}{c||ccc|ccc}
 	\hline
 	{RPN} &  			
	\multicolumn{3}{c|}{RPN - AP Score 3D} & \multicolumn{3}{c}{Refined - AP Score 3D} \\
 	&Easy & Moderate & Hard & Easy & Moderate & Hard\\
 	\hline\hline
 	Ours & 87.88 & 74.31 & 68.09 & 89.61 & 79.04 & 77.96 \\
 	SECOND v1.5 & 89.15 & 78.80 & 77.47 &  89.44 & 78.97 & 78.10 \\
 	PointRCNN & 89.10 & 78.71 & 77.79 &  89.58 & 79.09 & 78.06  \\
 	\hline
 	Ground Truth & 100.00 & 100.00 & 100.00 & 89.58 & 79.31 & 78.79\\
 %	\hline
 \end{tabular}
\end{center}
\caption{A comparison of different RPNs, refined by the LRN}
\label{tab:different_RPNs}
\end{table}

\paragraph{Domain Adaptation} 
We further investigate the option of a two-phase training procedure for our lightweight RPN.  As the patches occur at their original distance to the sensor, we train a detector that can operate on whole scenes. First, we pre-train by concentrating on the domain of patches only, then we switch to the domain of whole scenes. The pre-trained RPN surpasses the moderate 3D-scores of Voxelnet ($65.46$) after only one additional epoch.

\section{Conclusion}
We have proposed Patch Refinement, a two-stage model for 3D object detection from sparse LiDAR point clouds. 
We demonstrate that a modified Voxelnet is capable of highly accurate local bounding box regression and a simplified Voxelnet is an adequate choice for an RPN to complement such an LRN. As the LRN operates on local point cloud subsets only, it can refine the proposals of an arbitrary RPN on demand.
Further improvements regarding accuracy may be attainable by using a ground plane estimation algorithm and the integration of image information in the RPN stage.
{\small
\bibliographystyle{plain}
\bibliography{references}
}

\end{document}